\documentclass[letterpaper, 10 pt, journal, twoside]{IEEEtran}
\usepackage[caption=false,font=normalsize,labelfont=sf,textfont=sf]{subfig}
\usepackage{amsmath, amssymb, amsfonts} 
\usepackage{graphicx}
\usepackage{textcomp}
\usepackage{xcolor}
\usepackage{booktabs}
\usepackage{tabularx}
\usepackage{array}
\usepackage{xurl}
\usepackage{tikz}
\usepackage{pgfplots}
\pgfplotsset{compat=1.18}
\usepackage{multirow} 
\usepackage{algorithm}
\usepackage{algorithmic}
\usepackage{csquotes}
\usepackage{microtype}
\usepackage{cite}
\usepackage{hyperref}
\newcolumntype{Y}{>{\centering\arraybackslash}X}
\newcommand{\Com}[1]{\item[] $\triangleright$ \texttt{\small #1}}
\colorlet{mylightgray}{black!45}
\colorlet{mydarkgray}{black!100}

\begin{document}

\title{Multi-Robot Motion Planning from Vision and Language using Heat-inspired Diffusion}

\author{Jebeom Chae$^{1,*}$, Junwoo Chang$^{2,*}$, Seungho Yeom$^{2}$, Yujin Kim$^{2}$, and Jongeun Choi$^{1, 2,\dagger}$
\thanks{Manuscript received: December 4, 2025; Revised: February 23, 2026; Accepted: April 1, 2026.}%
\thanks{This paper was recommended for publication by Editor M. Ani Hsieh upon evaluation of the Associate Editor and Reviewers' comments. This work was supported by the National Research Foundation of Korea (NRF) grant funded by the Korea government (MSIT) (No.RS-2024-00344732). This work was also supported by the Korea Institute of Science and Technology (KIST) Institutional Program (Project No.2E33801-25-015). We also thank Hyunwoo Ryu for his insightful discussions.}

\thanks{$^{1}$Jebeom Chae and Jongeun Choi are with the Department of Artificial Intelligence, Yonsei University, Seoul, South Korea. 
{\tt\footnotesize \{jebeomchae, jongeunchoi\}@yonsei.ac.kr}}%
\thanks{$^*$Co-first authors. $^{\dagger}$Corresponding author.}%
\thanks{$^{2}$Junwoo Chang, Seungho Yeom, Yujin Kim, and Jongeun Choi are with the School of Mechanical Engineering, Yonsei University, Seoul, South Korea. 
{\tt\footnotesize \{junwoochang, duatmdgh3, djm06165, jongeunchoi\}@yonsei.ac.kr}}%
\thanks{Digital Object Identifier (DOI): see top of this page.}
}

\markboth{IEEE ROBOTICS AND AUTOMATION LETTERS. PREPRINT VERSION. ACCEPTED APRIL, 2026}%
{Chae \MakeLowercase{\textit{et al.}}: Multi-Robot Motion Planning from Vision and Language using Heat-inspired Diffusion}

\maketitle

\begin{abstract}
Diffusion models have recently emerged as powerful tools for robot motion planning by capturing the multi-modal distribution of feasible trajectories. 
However, their extension to multi-robot settings with flexible, language-conditioned task specifications remains limited. 
Furthermore, current diffusion-based approaches incur high computational cost during inference and struggle with generalization because they require explicit construction of environment representations and lack mechanisms for reasoning about geometric reachability. 
To address these limitations, we present Language-conditioned Heat-inspired Diffusion (LHD), an end-to-end vision-based framework that generates language-conditioned, collision-free trajectories. 
LHD integrates semantic priors from CLIP, a vision-language model (VLM), with a collision-avoiding diffusion kernel serving as a physical inductive bias that enables the planner to interpret language commands strictly within the reachable workspace.
This naturally handles out-of-distribution (OOD) scenarios---in terms of reachability---by guiding robots toward accessible alternatives that match the semantic intent, while eliminating the need for explicit obstacle information at inference time.
Extensive evaluations on diverse real-world-inspired maps, along with real-robot experiments, show that LHD consistently outperforms prior diffusion-based planners in success rate, while reducing planning latency. Project page is available at: \url{https://jebeom.github.io/lhd_project_page/}
\end{abstract}

\begin{IEEEkeywords}
Multi-Robot Systems, Diffusion Models, Motion and Path Planning, Collision Avoidance
\end{IEEEkeywords}

\section{Introduction}
\IEEEPARstart{M}{ulti-robot} motion planning (MRMP) is a fundamental problem in robotics, where teams of robots navigate shared environments while avoiding collisions. For real-world deployment in human-centric domains like automated warehouses, robots must be able to interpret and execute instructions from human operators, rather than relying on explicit goal coordinates. However, classical approaches are fundamentally limited in this context. They not only lack the ability to process language information but also struggle with scalability in complex continuous spaces. Search-based methods are often restricted to discrete domains \cite{sharon2015conflict, li2021eecbs}, sampling-based algorithms suffer from the curse of dimensionality \cite{lavalle1998rapidly, kavraki2002probabilistic, le2019multi}, and optimization-based approaches scale poorly with the number of robots due to expensive computational requirements \cite{augugliaro2012generation, park2020efficient}.

Learning-based methods have emerged as a promising alternative to handle these high-dimensional spaces. In particular, diffusion models have demonstrated remarkable success in single-robot motion planning \cite{janner2022planning}, effectively learning to satisfy hard constraints such as collision avoidance \cite{carvalho2023motion, feng2024ltldog}. Extending this capability to multi-robot settings, recent approaches have adopted hybrid strategies, such as combining a diffusion model with classical multi-agent path finding (MAPF) algorithms \cite{shaoul2024multi} or enforcing constraints via Lagrangian dual-based method \cite{liang2025simultaneous}. However, these methods suffer from significant challenges in terms of computational efficiency and generalization. Also, they lack the intrinsic capability to reason about geometric reachability, often failing when goals are physically obstructed.

\begin{figure*}[t]
\begin{center}
\includegraphics[width=1.0\linewidth]{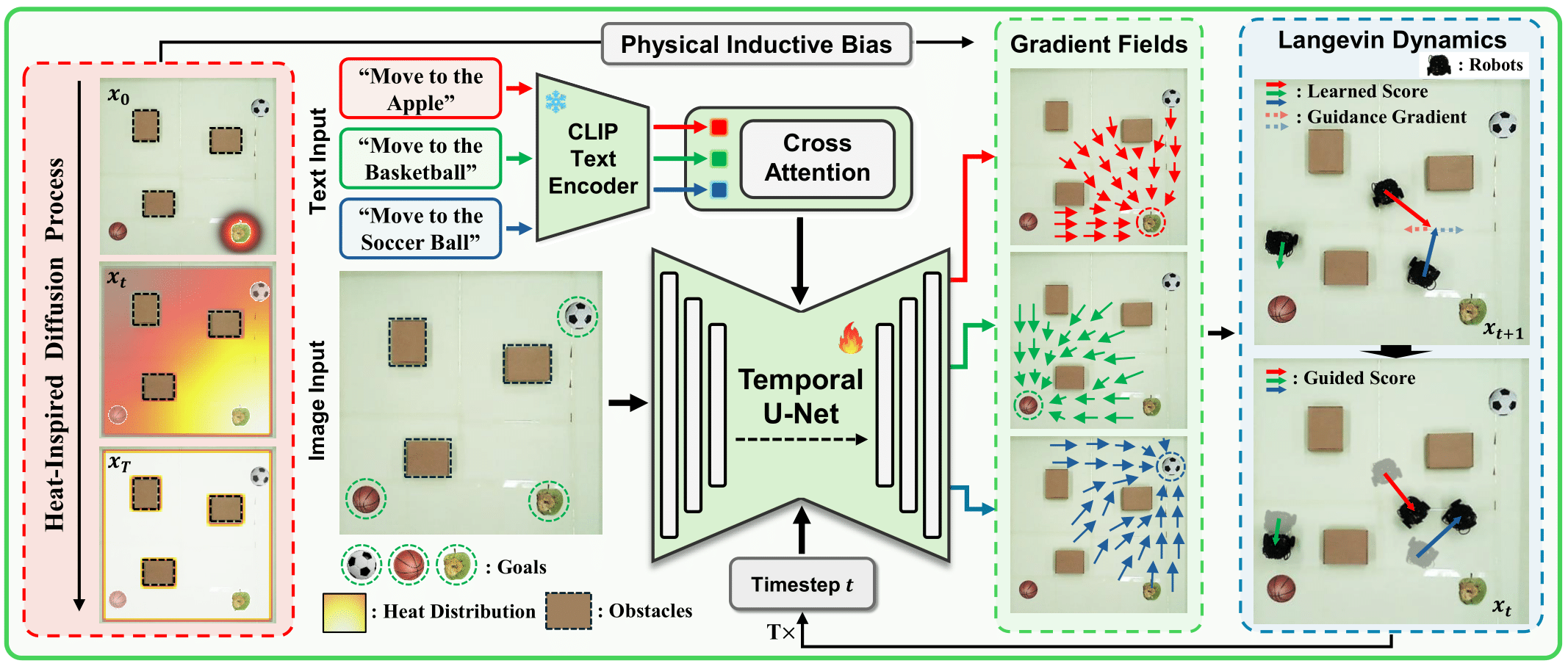}
\end{center}
\caption{\textbf{Overview of the LHD framework.} The model takes a raw RGB image and diffusion timestep t as inputs, conditioned on language instructions. A pre-trained CLIP text encoder \cite{radford2021learning} extracts fixed text embeddings, which are injected into the U-Net via cross-attention. The network outputs individual gradient fields, guiding each robot toward its respective goal while incorporating a heat-inspired physical inductive bias that inherently encodes reachability. During inference, Langevin dynamics iteratively sample the next state by aggregating these learned scores with inter-robot collision avoidance gradients, enabling safe multi-robot coordination.}
\label{architecture}
\vspace{-10pt}
\end{figure*}

To address these challenges, we propose Language-conditioned Heat-inspired Diffusion (LHD), an end-to-end vision-based multi-robot motion planning framework that integrates semantic priors from CLIP  \cite{radford2021learning}, a vision-language model (VLM), with a collision-avoiding diffusion kernel of DHD \cite{chang2023denoising}. 
This kernel serves as a physical inductive bias, which amortizes the cost of static obstacle avoidance into the training phase, thereby enabling the planner to interpret language commands strictly within the reachable workspace. 
Thus, it naturally resolves out-of-distribution (OOD) scenarios in terms of reachability by guiding robots toward accessible alternatives that maintain the semantic intent. 
Leveraging this implicit obstacle avoidance, we incorporate a simple coordination mechanism that enables multiple robots to safely share space during the reverse diffusion process. Consequently, LHD generates language-conditioned trajectories that satisfy both static obstacle and inter-robot safety constraints within practical planning times.
Fig.~\ref{architecture} provides an overview of the LHD framework.

Our contributions are as follows: 1) We introduce LHD, an end-to-end vision-language diffusion planner for multi-robot trajectory generation. 2) We propose a collision-avoiding diffusion kernel that grounds language instruction to reachable regions and amortizes static obstacle avoidance without explicit reconstruction. 3) We devise a lightweight inter-robot guidance term that enforces safety during sampling, and 4) extensive simulation and hardware results showing improved success rate, OOD generalization, and runtime over baselines.
\section{Related Works}
\textbf{Language Grounding in Robotics.}
Foundation models have significantly advanced language-conditioned robotics, ranging from LLM-based task decomposition \cite{ahn2022can} to VLA-based end-to-end control \cite{zitkovich2023rt, team2024octo}. Parallel to these generative approaches, VLMs, such as CLIP \cite{radford2021learning}, have been widely adopted for spatially grounding semantic concepts, enabling diverse applications in manipulation \cite{shridhar2022cliport} and navigation \cite{gadre2023cows}. To effectively integrate such semantic knowledge into generative models, recent diffusion models for text-to-image synthesis \cite{rombach2022high} have shown that cross-attention mechanisms can inject language conditioning at multiple stages while preserving generalization capability. Following this paradigm, recent diffusion-based robotic policies \cite{reuss2024multimodal, chen2025ac} adopt cross-attention conditioning with pre-trained encoders to integrate multi-modal observations. We extend this approach to multi-robot motion planning, generating language-conditioned gradient fields through cross-attention with a pre-trained CLIP text encoder.

\textbf{Multi-Robot Motion Planning.}
While classical approaches have established strong foundations, they often struggle with scalability in high-dimensional continuous spaces. 
Mean-field methods \cite{elamvazhuthi2020mean} address scalability by modeling swarm density evolution. One such method
\cite{zheng2021transporting} 
uses diffusion/heat equations for mean-field feedback control but remains limited to homogeneous teams with single global distribution targets.
Recently, diffusion models have emerged as a promising data-driven alternative. Leading approaches have adopted hybrid strategies to extend these models to multi-robot scenarios. 
MMD \cite{shaoul2024multi} combines diffusion models trained on single-robot data with MAPF, using diffusion guidance for obstacle avoidance and iterative replanning with constraints 
to resolve inter-robot collisions, while SMD \cite{liang2025simultaneous} enforces safety constraints by interleaving diffusion steps with Lagrangian dual projections. 
However, these methods exhibit limited robustness as they lack the capability to reason about geometric reachability, often failing in scenarios where designated goals are physically obstructed. 
In contrast, LHD addresses this limitation by embedding static collision avoidance into the training phase through its collision-avoiding diffusion kernel. 
This structural advantage enables practical planning times and robust performance even in OOD scenarios.
\section{Preliminary}
In this section, we define the problem statement and provide relevant background on motion planning with score-based diffusion models.
\subsection{Problem Statement}
We consider $N$ mobile robots operating in a shared 2D workspace $\mathcal{X} \subset \mathbb{R}^2$ with static obstacles $\mathcal{X}_{obs} \subset \mathcal{X}$. 
Robot $i$ has state $x^i_t\in \mathbb R^2$ at time $t$ and trajectory $\tau_i=\{x^i_t\}^T_{t=0}$. 
Given a top-down workspace observation $I\in \mathbb R^{H\times W\times C}$ aligned with $\mathcal X$ and a language instruction $l_i$ for each robot, we first determine a language-specified goal $g_i \in \mathcal X$ from $(I, l_i)$.
We then generate joint trajectories $\{\tau_i\}^N_{i=1}$ such that each robot reaches its goal (e.g., $\|x^i_T - g_i\|\leq \epsilon$ while satisfying: 1) $x^i_t\in \mathcal X ~ \backslash~ \mathcal X_{obs}$ for all $i,t$, and 2) $\|x^i_t-x^j_t\|\geq d_{\text{safe}}$ for all $i\neq j$ and $t$, where $d_{\text{safe}}$ is a user-specified minimum inter-robot safety margin. When multiple goal instances match the same language instruction, the agent should select a reachable instance (i.e., not isolated by obstacles).

\subsection{Motion Planning with Score-Based Diffusion Models}
\label{dm}
Score-based diffusion planning models robot trajectories as samples from $p_0(x)$. A Gaussian process defines $p_t(x)=\int p_0(x_0) q_t(x\mid x_0, \sigma^2_t I) dx_0,~q_t(x\mid x_0)=\mathcal N(x;x_0, \sigma^2_t I)$, and a score network $s_\theta(x,t)$ is trained via score matching \cite{hyvarinen2005estimation, vincent2011connection} to approximate $\nabla_x \log p_t(x)$. At inference, sampling starts from $x_T\sim \mathcal{N}(0, \sigma^2_T I)$ and applies reverse-time updates driven by the learned score to obtain a trajectory \cite{grenander1994representations}. See \cite{song2019generative, song2020score} for details. This framework is closely related to  DDPM \cite{ho2020denoising}, where the noise-prediction objective effectively minimizes a weighted score matching loss to parameterize the score. While recent works have successfully applied diffusion to planning \cite{janner2022planning, carvalho2023motion} and closed-loop control \cite{ chi2025diffusion}, we specifically employ a score-based diffusion planner integrated with a frozen text encoder for language-conditioned goal specification, supplemented by a collision-aware prior that biases sampling toward safe and reachable joint trajectories.

\section{Method}
We present Language-conditioned Heat-inspired Diffusion (LHD), an approach for multi-robot motion planning that generates language-conditioned, collision-free trajectories from a raw RGB image. 
By integrating CLIP-based semantic priors with physical priors from heat transfer, LHD physically grounds language instructions within the reachable workspace, while amortizing static obstacle avoidance into the training phase.
First, we describe how collision constraints with static obstacles are embedded into the forward diffusion process through collision-avoiding diffusion kernel (Sec.~\ref{sec:heat_kernel}). Second, we show that trajectories emerge from the reverse diffusion process without trajectory-level supervision (Sec.~\ref{sec:score_learning}). Third, we discuss how language instructions are integrated into the model for semantic goal specification (Sec.~\ref{sec:language_cond}). Finally, we address inter-robot collision avoidance during inference via simple distance-based guidance (Sec.~\ref{sec:guidance}).

\subsection{Collision-Avoiding Diffusion Kernel}
\label{sec:heat_kernel}
Traditional diffusion-based motion planners rely on Gaussian kernels which lack explicit collision-avoidance mechanisms. Consequently, recent approaches require measurement of the distance from the obstacle or auxiliary inputs to incorporate obstacle information. To overcome this problem, we adopt the collision-avoiding diffusion kernel introduced by \cite{chang2023denoising} which embeds collision constraints directly into the diffusion kernel via the heat equation: 
\begin{equation}
    \label{heat}
    \frac{\partial u}{\partial t} = \nabla\cdot (K(x)\nabla u),
\end{equation}
where $u$ denotes the heat distribution and $K(x)$ is the thermal conductivity field. They interpret the heat distribution $u$ governed by Eq.\ref{heat} as the perturbed distribution $p_t(x)$ used in the diffusion process. By modeling obstacles as perfect insulators that block heat flow, the resulting perturbed distribution inherently excludes unreachable regions. This physically-grounded formulation guides the model to learn collision-avoidance behavior, mimicking the way heat diffuses only through traversable space. In practice, we numerically solve Eq.~\ref{heat} using the Finite Difference Method (FDM) to obtain the ground-truth heat distribution $u_t$, and compute its log-gradient $\nabla \log u_t$ to serve as the training target (refer to line~\ref{alg:lhd:line6} in Alg. \ref{alg:lhd}).

\subsection{Trajectory Generation from Heat-inspired Diffusion}
\label{sec:score_learning}
Unlike traditional trajectory diffusion models that require full path demonstrations, our approach obtains trajectories as a natural byproduct of the diffusion process itself. Given a goal $\mathbf{x}_0$, we compute heat distributions $u_t$ at various diffusion times by solving Eq.~\ref{heat} with heat sources at $\mathbf{x}_0$. These heat distributions define gradient fields that encode not just the goal location, but the entire geometry of how heat diffuses from goals through free space while avoiding obstacles. 

This formulation provides a significant efficiency advantage during inference. In standard diffusion models, intermediate denoising steps serve merely as a computational mechanism to reach the final output, with only the final generated state being utilized. In contrast, our approach treats every intermediate denoised state $\mathbf{x}_t$ as a meaningful waypoint that represents a collision-free configuration progressively approaching the goal. Thus, a single reverse diffusion process simultaneously produces both the final goal state and a complete executable trajectory $\{\mathbf{x}_T, \mathbf{x}_{T-1}, ..., \mathbf{x}_0\}$.

\begin{algorithm}[t]
\caption{Language-conditioned Heat-inspired Diffusion}
\label{alg:lhd}
\begin{algorithmic}[1]
\small

\item[] \textbf{------TRAINING------} 
\item[] \textbf{Input:} Top-down view images $\mathcal{Y}$, Obstacle masks $\mathcal{O}$, Goal positions $\mathcal{G}$,  Language Instructions $\mathcal{L}$, Diffusion model $\mathbf{s}_\theta$, Learning rate $\alpha$, Total diffusion timesteps $T$

\WHILE{training is not finished}
    \Com{sample a batch of training data}
    \STATE ${y} \sim \mathcal{Y}$, $y_\text{obs} \sim \ \mathcal{O}$, $\mathbf{x}_0 \sim \ \mathcal{G}$, $\ell \sim \mathcal{L}$, $t \sim \mathcal{U}(1, T)$
    \Com{encode text instruction}
    \STATE$ \mathbf{z} = \text{CLIP}_{\text{frozen}}(\ell)$
    \Com{compute target score}
    \STATE $\nabla \log u_t, \mathbf{x}_t \sim \text{ForwardHeat}(\mathbf{x}_0, y_{\text{obs}}, t)$  \label{alg:lhd:line6}
    \Com{predict gradient field from network}
    \STATE $\mathbf{S}_t = s_\theta(y, t, \mathbf{z})$ \label{alg:lhd:line7}
    \Com{query score at perturbed position}
    \STATE $\hat{\mathbf{s}} = \text{BilinearInterp}(\mathbf{S}_t, \mathbf{x}_t)$ \label{alg:lhd:line8}
    \Com{compute the score matching loss}
    \STATE $\mathcal{L}(\theta) = \| \nabla \log u_t - \hat{\mathbf{s}} \|^2_2$ \label{alg:lhd:line9}
    \Com{gradient update}
    \STATE $\theta = \theta - \alpha \nabla_\theta \mathcal{L}(\theta)$ 
\ENDWHILE

\item[] \textbf{------INFERENCE------}
\item[] \textbf{Input:} Pre-trained diffusion model $\mathbf{s}_\theta$, Top-down view image $y$, Language Instruction $\ell$, Number of robots $N$, Annealing steps $K$
\item[] \textit{\small (Superscript $(n)$ denotes batch processing over $N$ robots) \vspace{3pt}}

\STATE $\mathbf{z}^{(n)} \gets \text{CLIP}_{\text{frozen}}(\ell^{(n)})$ 
\Com{sample initial positions from free space\vspace{3pt}}
\STATE $\mathbf{x}_T^{(n)} \sim \mathcal{U}(\mathcal{X}_{\text{free}})$
\FOR{$t = T, \ldots, 1$}
    \STATE $\mathbf{S}_t^{(n)} = s_\theta(y, t, \mathbf{z}^{(n)})$
    \FOR{$k = 1, \ldots, K$}
        \STATE $\mathbf{s}^{(n)}_t = \text{BilinearInterp}(\mathbf{S}_t^{(n)}, \mathbf{x}_t^{(n)})$ 
        \Com{perform Langevin Dynamics\vspace{3pt}}
        \STATE $\mathbf{x}_{t-1}^{(n)} = \mathbf{x}_t^{(n)} + 0.5 \alpha_t^2 [\mathbf{s}^{(n)}_t + \beta \cdot \nabla c_{\text{int}}^{(n)}] + \alpha_t \boldsymbol{\epsilon}$  \vspace{6pt} \label{alg:lhd:line16}
    \ENDFOR
\ENDFOR

\item[] \textbf{Output:} Robots' trajectories $\{(\mathbf{x}_T^{(n)}, \ldots, \mathbf{x}_1^{(n)}, \mathbf{x}_0^{(n)})\}_{n=1}^N$

\end{algorithmic}
\end{algorithm}
\subsection{Language Conditioning for Semantic Goal Specification}
\label{sec:language_cond}
We extend this collision-avoiding planner with language conditioning to enable flexible, task-specific control through a frozen CLIP text encoder \cite{radford2021learning}. Given a language instruction, we extract a text embedding $\mathbf{z}$ and inject it into every U-Net block via cross-attention.
This architectural choice allows the score model to dynamically attend to relevant linguistic features while processing spatial information.

We train a score model using the temporal U-Net backbone \cite{ho2020denoising} that takes a top-down view image $y$ as input. Since the model outputs a gradient field $\mathbf{S}_t$ on a discrete spatial grid, we use bilinear interpolation to evaluate the score $s_\theta(\mathbf{x}_t, t, y,\mathbf{z})$ at arbitrary continuous positions $\mathbf{x}_t$ (cf. lines~\ref{alg:lhd:line7}--\ref{alg:lhd:line8}) in Algorithm~\ref{alg:lhd}. The training objective combines the collision-avoiding diffusion process with language conditioning via denoising score matching:
\begin{equation}
\begin{split}
    \min_\theta \mathbb{E}_{t, y, y_{obs}, \mathbf{z}, \mathbf{x}_0, \mathbf{x}_t} \Big[ & \lambda(t) \|s_\theta(\mathbf{x}_t, t, y, \mathbf{z}) \\
    & - \nabla_{\mathbf{x}_t} \log p(\mathbf{x}_t \mid \mathbf{x}_0, y_{obs}) \|^2_2 \Big],
\end{split}
\end{equation}
where $y_{obs}$ represents the corresponding obstacle mask. The diffusion time step $t$ is sampled uniformly from the interval $[0, T]$, while the goal state $x_0$ is drawn from the distribution of reachable 
goals $p_0(x_0)$. The perturbed state $x_t$ is then generated through the forward diffusion process. The time-dependent weighting function $\lambda(t)$, originally proposed in \cite{song2020score}, is used to scale the loss at each time step. This loss is computed following the procedure in line \ref{alg:lhd:line9} of Algorithm~\ref{alg:lhd}.

\begin{figure*}[t]
    \centering \includegraphics[width=0.97\linewidth]{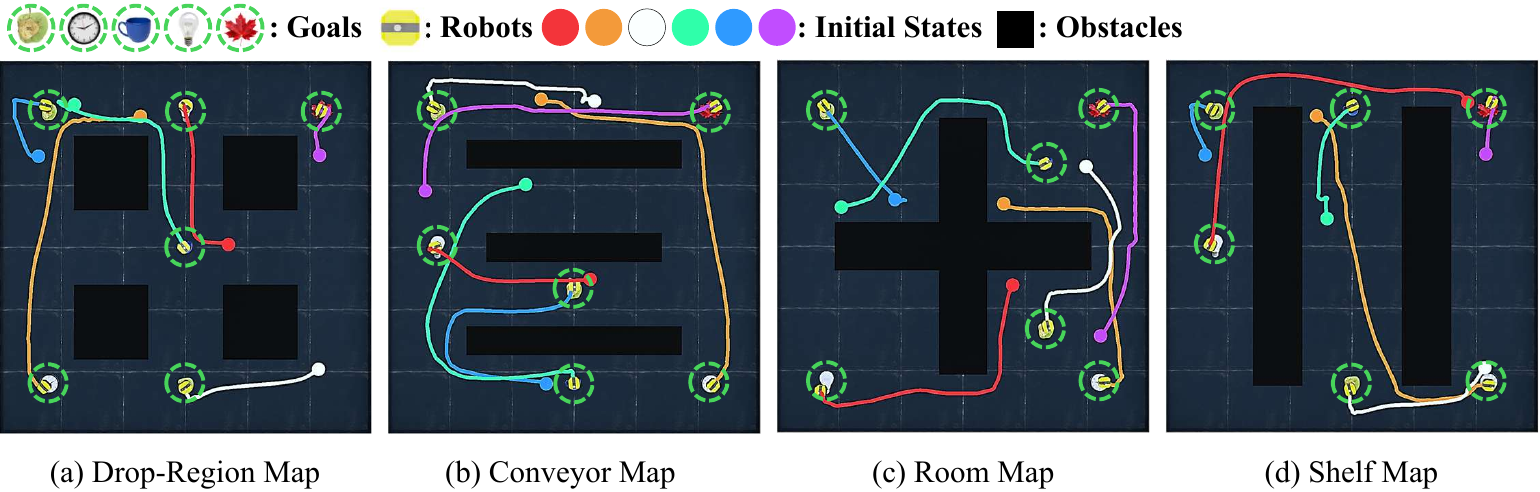}
    \caption{\textbf{Qualitative results of our proposed method.} The figures demonstrate language-conditioned, collision-free trajectories across four real-world-inspired environments: \textbf{(a)} Drop-Region, \textbf{(b)} Conveyor, \textbf{(c)} Room, and \textbf{(d)} Shelf maps. Colored dots indicate the start positions of the robots, and the corresponding colored lines represent their trajectories to the goals.}
    \label{fig:simulation}
    \vspace{-15pt}
\end{figure*}

\subsection{Guided Sampling for Inter-Robot Collision Avoidance}
\label{sec:guidance}

The collision-avoiding diffusion kernel from Sec.~\ref{sec:heat_kernel} ensures collision-free paths with respect to static obstacles. However, multi-robot motion planning additionally requires avoiding collisions between robots at every point along their trajectories. Since robot positions change dynamically, we cannot pre-encode these constraints in the diffusion kernel.

Instead, we address this through guided sampling during the reverse diffusion process. 
At each diffusion step $t$, each robot receives goal-directed guidance from its own gradient field computed based on its language instruction. We then perform $K$ annealing steps, where we augment the Langevin dynamics with a gradient of an inter-robot collision cost (line~\ref{alg:lhd:line16} in Algorithm~\ref{alg:lhd}):
\begin{equation}
    \label{eq:langevin}
    \mathbf{x}_{t-1} = \mathbf{x}_t + 0.5~\alpha^2_t \left[s_\theta(\mathbf{x}_t, t, y, \mathbf{z}) + \beta ~\nabla_{\mathbf{x}_t} c_{\text{int}}(\mathbf{x}_t)\right] + \alpha_t \boldsymbol{\epsilon}\textcolor{blue}{,}
\end{equation}
where $\boldsymbol{\epsilon}\sim \mathcal{N}(0, I)$, $\alpha_t \propto \sigma_t$ follows the forward noise schedule, and $\beta$ controls the guidance strength. The inter-robot collision cost is defined as:
\begin{equation}
c_{\text{int}}(\mathbf{x}_t) = \sum_{i<j} \max \left( 0, -\log \left( \frac{\|\mathbf{x}^i_t - \mathbf{x}^j_t\|_2}{d_{\text{margin}}} \right) \right),
\end{equation}
which penalizes robot pairs when their distance falls below an interaction threshold $d_{\text{margin}}$ at diffusion step $t$. Here, $d_{\text{margin}}$ is set slightly larger than the safety distance $d_{\text{safe}}$ to provide a safety margin. By applying this guidance at every denoising step, the model generates coordinated, collision-free trajectories where robots maintain safe distances throughout their entire paths while respecting their individual language-specified goals. The complete training and inference procedures are detailed in Algorithm~\ref{alg:lhd}.

\section{Experiments}
We evaluate LHD on multi-robot motion planning tasks to demonstrate: (i) its performance against state-based approaches requiring explicit obstacle representations and auxiliary goal extraction, (ii) generalization to OOD scenarios in terms of reachability and visual robustness, (iii) scalability to diverse environments with varying numbers of robots in both simulation and the real-world.

\subsection{Experimental Settings}
\textbf{Maps.} We validate LHD on four benchmark maps adapted from prior work \cite{shaoul2024multi, liang2025simultaneous}, representing diverse real-world planning challenges:
\begin{itemize}
    \item \textbf{Drop-Region map} features designated pickup and delivery zones with open navigation areas, testing coordination in structured warehouse-like environments.
    \item \textbf{Conveyor map} simulates constrained corridors around conveyor belts, requiring robots to navigate through narrow passages while avoiding static obstacles.
    \item \textbf{Room map} contains multiple rooms connected by doorways, restricting simultaneous entry and requiring careful scheduling to prevent congestion.
    \item \textbf{Shelf map} models warehouse storage layouts with tight aisles between shelves, demanding precise multi-robot coordination in confined spaces.
\end{itemize}

\textbf{Task Specification.} The core task requires each robot to move from its initial position to its assigned goal region, where the target destination is inferred directly from a raw RGB image and natural language instructions. Start and goal positions are randomly sampled within obstacle-free regions of each map. For each map, we conduct experiments with 3, 6, and 9 robots, generating 10 test cases per configuration across 12 different map variants with varying obstacle configurations, including different obstacle sizes and positions.

\textbf{Evaluation Metrics.} We assess LHD using three primary metrics. \underline{Success Rate} indicates the proportion of test cases solved without collisions and reaching the target regions within the time limit (180~seconds). \underline{Planning Time} measures the computational efficiency, reflecting the practical applicability of the approach.
\underline{Path Length} measures the average travel distance per robot within a physical workspace scaled to $18 \times 18$\,m, reflecting efficiency.

\textbf{Implementation.} 
We implemented our method and all baselines in Python. In our experiments, the size of each local map is normalized to $2 \times 2$~units. Our model uses the frozen CLIP ViT-B/32 text encoder for language conditioning and a diffusion process with 20 denoising steps, trained using the Adam optimizer with learning rate $10^{-4}$ and batch size 48. Our quantitative benchmarks were conducted on a workstation equipped with an Intel Core i9-13900KF CPU and an NVIDIA RTX 4090 GPU. For real-world validation, planning was performed on a separate PC equipped with an AMD Ryzen 9 9900X CPU and an NVIDIA RTX 5060 Ti GPU.

\textbf{Baselines.} 
We compare LHD against representative methods from both classical search-based and learning-based approaches. For search-based methods, we evaluate Explicit Estimation CBS (EECBS) \cite{li2021eecbs}, a state-of-the-art bounded-suboptimal multi-agent path finding (MAPF) algorithm that operates on discretized grids. For learning-based methods, we benchmark whether an off-the-shelf language goal specification module, combined with a open-loop diffusion planner, can solve the task without our collision-aware prior. Specifically, we compare against four baselines: 1) Standard Diffusion Model (DM) \cite{ho2020denoising} trained on multi-robot trajectories; 2) Motion Planning Diffusion (MPD) \cite{carvalho2023motion}, a state-of-the-art single-robot motion planning diffusion model adapted to multi-robot settings; 3) Multi-Robot Multi-Model Planning Diffusion (MMD) \cite{shaoul2024multi}, which coordinates single-robot diffusion models through conflict-based search with iterative replanning; and 4) Simultaneous MRMP Diffusion (SMD) \cite{liang2025simultaneous}, which integrates Lagrangian dual-based constrained optimization directly into the diffusion sampling process. 
While we also explored recent multi-agent reinforcement learning (MARL) algorithms, they struggled to satisfy strict collision constraints in our complex environments.
We utilize the pre-trained checkpoints provided in the official repositories of MMD \cite{shaoul2024multi} and SMD \cite{liang2025simultaneous} for all learning-based baselines. 
All baselines are augmented with Lang-SAM \cite{medeiros2023langsam}, a vision-language grounding model that extracts goal coordinates from visual inputs and task prompts. 
Note that planning times reported in Table~\ref{tab:main_result} for baseline methods include the Lang-SAM inference overhead ($\approx$0.1s), while LHD's planning times reflect end-to-end performance without additional preprocessing.

\begin{table*}[t]
    \centering
    \caption{\textbf{Quantitative comparison of our method against baselines across four real-world-inspired maps.} $n$ denotes the number of robots. The reported metrics are Success Rate $S$, Average Planning Time $T$ (s), and Path Length $L$ (m). Results are marked with a hyphen (-) for cases with timeouts (180s) or planning fail. Gray shades denote configurations with $S < 50\%$, while bold highlights the best performance among models achieving $S \geq 80\%$.}
    \label{tab:main_result}
    \small
    \setlength{\tabcolsep}{6.0pt} 
    \renewcommand{\arraystretch}{0.93}
    \begin{minipage}[t]{0.49\textwidth}
    \centering
    \begin{tabular}{c c | c c c c c c}
        \toprule
        \multicolumn{8}{c}{Drop-Region Maps} \\
        \midrule
        n & Metric & EECBS & DM & MPD & MMD & SMD & \textbf{Ours} \\ 
        \midrule
        \multirow{3}{*}{3} 
           & $S \uparrow$     & \textbf{1}     & \textcolor{mylightgray}{0.05} & 0.97     &  0.95    &  \textbf{1}     & \textbf{1} \\
           & $T \downarrow$ & \textbf{0.10}     & \textcolor{mylightgray}{0.17} & 1.01     &  1.82    & 41.33     & 0.24 \\
           & $L \downarrow$ & 8.98 & \textcolor{mylightgray}{5.05} & 9.12 & 8.25 & \textbf{7.95} & 8.90 \\
        \midrule
        \multirow{3}{*}{6} 
           & $S \uparrow$     & \textbf{1}    & \textcolor{mylightgray}{0}     & \textcolor{mylightgray}{0} &  0.86    &  \textcolor{mylightgray}{0.43}    & \textbf{1} \\
           & $T \downarrow$ & \textbf{0.11}   & \textcolor{mylightgray}{0.17}     & \textcolor{mylightgray}{0.91} & 3.67     & \textcolor{mylightgray}{136.93}     & 0.37 \\
           & $L \downarrow$ & 9.46 & \textcolor{mylightgray}{-} & \textcolor{mylightgray}{-} & \textbf{8.52} & \textcolor{mylightgray}{14.48} & 10.07 \\
        \midrule
        \multirow{3}{*}{9} 
           & $S \uparrow$     & \textbf{1}     & \textcolor{mylightgray}{0}     & \textcolor{mylightgray}{0}     &  \textcolor{mydarkgray}{0.73}    & -    & \textbf{1} \\
           & $T \downarrow$ & \textbf{0.11}     & \textcolor{mylightgray}{0.18}     & \textcolor{mylightgray}{0.92}     & \textcolor{mydarkgray}{5.50} & -   & 0.47 \\
           & $L \downarrow$ & \textbf{9.51} & \textcolor{mylightgray}{-} & \textcolor{mylightgray}{-} & \textcolor{mydarkgray}{8.59} & - & 10.28 \\
        \bottomrule
    \end{tabular}
    \end{minipage}
    \hfill
    \begin{minipage}[t]{0.49\textwidth}
    \centering
    \begin{tabular}{c c | c c c c c c}
        \toprule
        \multicolumn{8}{c}{Conveyor Maps} \\
        \midrule
        n & Metric & EECBS & DM & MPD & MMD & SMD & \textbf{Ours} \\ 
        \midrule
        \multirow{3}{*}{3} 
           & $S \uparrow$     & \textbf{1}     & \textcolor{mylightgray}{0.11} & \textcolor{mylightgray}{0.15} & \textcolor{mydarkgray}{0.69} &  \textcolor{mylightgray}{0.18}    & \textbf{1} \\
           & $T \downarrow$ & \textbf{0.10}    & \textcolor{mylightgray}{0.17} & \textcolor{mylightgray}{1.00} & \textcolor{mydarkgray}{1.75} & \textcolor{mylightgray}{35.46} & 0.23 \\
           & $L \downarrow$ & 11.38 & \textcolor{mylightgray}{5.67} & \textcolor{mylightgray}{6.93} & \textcolor{mydarkgray}{9.40} & \textcolor{mylightgray}{6.50} & \textbf{10.32} \\
        \midrule
        \multirow{3}{*}{6} 
           & $S \uparrow$     &    \textbf{1}    &  \textcolor{mylightgray}{0}  & \textcolor{mylightgray}{0}     & \textcolor{mylightgray}{0.39} &  -      & \textbf{1} \\
           & $T \downarrow$ & \textbf{0.11}     & \textcolor{mylightgray}{0.18}    & \textcolor{mylightgray}{0.90} & \textcolor{mylightgray}{3.58} & -      & 0.37 \\
           & $L \downarrow$ & \textbf{10.65} & \textcolor{mylightgray}{-} & \textcolor{mylightgray}{-} & \textcolor{mylightgray}{9.07} & - & 11.04 \\
        \midrule
        \multirow{3}{*}{9}  
           & $S \uparrow$     & \textbf{1}     & \textcolor{mylightgray}{0}     & \textcolor{mylightgray}{0}    & \textcolor{mylightgray}{0.14} &  -      & \textbf{1} \\
           & $T \downarrow$ & \textbf{0.11}  & \textcolor{mylightgray}{0.18}    & \textcolor{mylightgray}{0.93}     & \textcolor{mylightgray}{5.60} & -      & 0.49 \\
           & $L \downarrow$ & \textbf{10.85} & \textcolor{mylightgray}{-} & \textcolor{mylightgray}{-} & \textcolor{mylightgray}{8.93} & - & 11.40 \\
        \bottomrule
    \end{tabular}
    \end{minipage}
    
    \vspace{0.10cm} 
    
    \begin{minipage}[t]{0.49\textwidth}
    \centering
    \begin{tabular}{c c | c c c c c c}
        \toprule
        \multicolumn{8}{c}{Room Maps} \\
        \midrule
        n & Metric & EECBS & DM & MPD & MMD & SMD & \textbf{Ours} \\ 
        \midrule
        \multirow{3}{*}{3} 
           & $S \uparrow$     & \textbf{1}     & \textcolor{mylightgray}{0.07} & \textcolor{mylightgray}{0.13} & \textcolor{mydarkgray}{0.53} &  \textcolor{mylightgray}{0.35}      & \textbf{1} \\
           & $T \downarrow$ & \textbf{0.10}     & \textcolor{mylightgray}{0.17} & \textcolor{mylightgray}{1.01} & \textcolor{mydarkgray}{1.76} & \textcolor{mylightgray}{61.87}      & 0.24 \\
           & $L \downarrow$ & 12.58 & \textcolor{mylightgray}{3.41} & \textcolor{mylightgray}{6.32} & \textcolor{mydarkgray}{8.89} & \textcolor{mylightgray}{7.37} & \textbf{11.16} \\
        \midrule
        \multirow{3}{*}{6} 
           & $S \uparrow$     & \textbf{1}     & \textcolor{mylightgray}{0}    & \textcolor{mylightgray}{0}    & \textcolor{mylightgray}{0.26} & \textcolor{mylightgray}{0.01}      & 0.99 \\
           & $T \downarrow$ & \textbf{0.11}     & \textcolor{mylightgray}{0.17}     & \textcolor{mylightgray}{0.92} & \textcolor{mylightgray}{3.80} & \textcolor{mylightgray}{158.06}  & 0.37 \\
           & $L \downarrow$ & 12.73 & \textcolor{mylightgray}{-} & \textcolor{mylightgray}{-} & \textcolor{mylightgray}{9.29} & \textcolor{mylightgray}{12.33} & \textbf{12.16} \\
        \midrule
        \multirow{3}{*}{9} 
           & $S \uparrow$     & \textbf{1}     & \textcolor{mylightgray}{0}     & \textcolor{mylightgray}{0}     & \textcolor{mylightgray}{0.11} & -      & 0.99 \\
           & $T \downarrow$ & \textbf{0.12}     & \textcolor{mylightgray}{0.18}     & \textcolor{mylightgray}{0.93}     & \textcolor{mylightgray}{6.02} & -       & 0.47 \\
           & $L \downarrow$ & \textbf{12.98} & \textcolor{mylightgray}{-} & \textcolor{mylightgray}{-} & \textcolor{mylightgray}{9.21} & - & 13.03 \\
        \bottomrule
    \end{tabular}
    \end{minipage}
    \hfill
    \begin{minipage}[t]{0.49\textwidth}
    \centering
    \begin{tabular}{c c | c c c c c c}
        \toprule
        \multicolumn{8}{c}{Shelf Maps} \\
        \midrule
        n & Metric & EECBS & DM & MPD & MMD & SMD & \textbf{Ours} \\ 
        \midrule
        \multirow{3}{*}{3} 
           & $S \uparrow$     & \textbf{1}     & \textcolor{mylightgray}{0.05} & \textcolor{mylightgray}{0.10}     &  \textcolor{mylightgray}{0.33}      & \textcolor{mylightgray}{0.12}    & \textbf{1} \\
           & $T \downarrow$ & \textbf{0.10}     & \textcolor{mylightgray}{0.17} & \textcolor{mylightgray}{0.99}       & \textcolor{mylightgray}{1.74}      &  \textcolor{mylightgray}{46.25}  & 0.24  \\
           & $L \downarrow$ & 11.81 & \textcolor{mylightgray}{4.78} & \textcolor{mylightgray}{6.12} & \textcolor{mylightgray}{8.13} & \textcolor{mylightgray}{6.67} & \textbf{10.96} \\
        \midrule
        \multirow{3}{*}{6} 
           & $S \uparrow$     & \textbf{1}     & \textcolor{mylightgray}{0}     & \textcolor{mylightgray}{0} &  \textcolor{mylightgray}{0.13}      &  -      & 0.98 \\
           & $T \downarrow$ & \textbf{0.11}     & \textcolor{mylightgray}{0.18}     & \textcolor{mylightgray}{0.91} &  \textcolor{mylightgray}{3.50}      & -      & 0.37 \\
           & $L \downarrow$ & 12.14 & \textcolor{mylightgray}{-} & \textcolor{mylightgray}{-} & \textcolor{mylightgray}{9.08} & - & \textbf{11.79} \\
        \midrule
        \multirow{3}{*}{9} 
           & $S \uparrow$     & \textbf{1}     & \textcolor{mylightgray}{0}     & \textcolor{mylightgray}{0}     & \textcolor{mylightgray}{0.04}      & -      & 0.98 \\
           & $T \downarrow$ & \textbf{0.11}     & \textcolor{mylightgray}{0.18}     & \textcolor{mylightgray}{0.93}  & \textcolor{mylightgray}{5.90} & -      & 0.47 \\
           & $L \downarrow$ & \textbf{12.16} & \textcolor{mylightgray}{-} & \textcolor{mylightgray}{-} & \textcolor{mylightgray}{9.71} & - & 12.49 \\
        \bottomrule
    \end{tabular}
    \end{minipage}
    
    \vspace{-15pt}
\end{table*}

\begin{table}[h]
    \centering
    \footnotesize
    \setlength{\tabcolsep}{6pt} 
    \caption{\textbf{Scalability evaluation of LHD with 15--30 robots across four maps.} The reported metrics are Success Rate ($S$) and Average Planning Time ($T$) in seconds.}
    \label{tab:scalability}
    \begin{tabularx}{\columnwidth}{@{} cc *{4}{Y} @{}}
        \toprule
        \textbf{\# Robots} & \textbf{Metric} & \textbf{Drop} & \textbf{Conveyor} & \textbf{Room} & \textbf{Shelf} \\
        \midrule
        \multirow{2}{*}{15} & $S \uparrow$   & 0.99 & 0.98 & 0.97 & 0.95 \\
                            & $T \downarrow$ & 0.76 & 0.76 & 0.76 & 0.76 \\
        \midrule
        \multirow{2}{*}{20} & $S \uparrow$   & 0.98 & 0.95 & 0.87 & 0.91 \\
                            & $T \downarrow$ & 1.02 & 1.02 & 1.02 & 1.02 \\
        \midrule
        \multirow{2}{*}{25} & $S \uparrow$   & 0.99 & 0.94 & 0.72 & 0.86 \\
                            & $T \downarrow$ & 1.26 & 1.25 & 1.24 & 1.25 \\
        \midrule
        \multirow{2}{*}{30} & $S \uparrow$   & 0.93 & 0.86 & 0.53 & 0.72 \\
                            & $T \downarrow$ & 1.48 & 1.48 & 1.45 & 1.48 \\
        \bottomrule
    \end{tabularx}
\end{table}
\subsection{Comparison of Methods}
\label{comparison}
We now compare LHD against the described baselines. The full quantitative results across all metrics and scenarios are summarized in Table \ref{tab:main_result}.

\textbf{Explicit Estimation CBS (EECBS).} While EECBS demonstrates high success rates and fast planning times across all tested scenarios, consistently solving problems with up to 9 robots, it is fundamentally constrained by its reliance on discrete grid spaces ($32 \times 32$ in our implementation). This spatial discretization inherently limits trajectory smoothness, producing grid-aligned paths that lack the kinematic feasibility required for direct execution. Thus, additional post-processing is required to convert these paths into executable trajectories.

\textbf{Standard Diffusion Models (DM).} Despite being trained on multi-robot trajectory data, DM exhibits poor performance, achieving success rates below 12\% for 3~robots and failing completely (0\%) with 6 or more robots across all environments.
This breakdown stems from the difficulty of learning multi-robot coordination in high-dimensional joint spaces, where the model fails to capture effective coordination patterns as team size increases.

\textbf{Motion Planning Diffusion (MPD).} MPD shows limited applicability, performing effectively only in simple, low-constrained settings. 
While it achieves a high success rate of 97\% with 3~robots in the Drop-Region map, its performance degrades precipitously in complex environments. 
Furthermore, it exhibits complete failure with 6 or more robots across all scenarios, despite attempting to correct trajectories via distance-based guidance.
This indicates that the learned prior in MPD suffers from the same high-dimensional joint distribution problem as DM.

\textbf{Multi-Robot Multi-Model Planning
Diffusion (MMD).} While MMD improves over MPD by utilizing MAPF logic, its effectiveness is largely confined to less constrained environments. In the Drop-Region maps, MMD maintains a relatively high success rate of 73\% with 9 robots, significantly outperforming MPD. 
However, success rates plummet to 11\% and 4\% in the Room and Shelf maps with 9 robots, respectively.
This failure stems from MMD's reliance on distance-based guidance to simultaneously handle both static obstacles and inter-robot collisions during inference. In narrow passages, enforcing these constraints restricts the feasible solution space, causing the planner to fail in generating a valid trajectory.

\textbf{Simultaneous MRMP Diffusion (SMD).} Unlike prior methods that rely on distance guidance for collision avoidance, SMD employs a Lagrangian dual-based optimization framework to rigorously enforce collision constraints. In relatively simple scenarios, such as the Drop-Region maps with 3~robots, this approach proves reliable, achieving a 100\% success rate with an average planning time of 41~seconds. 
However, the computational burden grows rapidly with the number of robots and map complexity, frequently causing planning times to exceed the 180-second cutoff in these scenarios.
Consequently, successful outcomes were largely confined to instances with shorter start-goal distances, rendering the method impractical for time-sensitive applications.

\textbf{{Language-conditioned Heat-inspired Diffusion (LHD).}} LHD consistently outperforms learning-based baselines in both success rate and planning time. For instance, in the Drop-Region maps with 6~robots, LHD achieves a 100\% success rate in 0.37~seconds. In contrast, SMD attains only a 43\% success rate while averaging 136.93~seconds, representing a speedup of over $300\times$, even excluding timed-out failures.
Although EECBS, which is a grid-based discrete search planner, achieves the best success rate and planning time, LHD yields average improvement factors of $5.18 \times 10^3$ in Smoothness, $3.42$ in Energy, $12.34$ in Minimum Distance to Obstacles, and $1.73$ in Minimum Distance to Robots.
However, as shown in Table \ref{tab:main_result}, LHD occasionally produces relatively longer path lengths.
This result stems from the inherent characteristic of heat propagation dynamics, where heat initially accumulates at the insulator boundaries. 
Consequently, trajectories may deviate toward walls or obstacles before converging to the goal, as illustrated by the blue agent’s path in Fig. \ref{fig:simulation}(a).
Such behavior reflects a fundamental, yet tunable, safety-efficiency trade-off in our approach. 
While reducing the score's influence during sampling can improve path efficiency, it may increase collision risks, thereby reducing overall safety.

\textbf{Ablation Study.} We conducted a brief ablation study on LHD in the 6-robot Drop-Region scenario. Removing the collision-avoiding diffusion kernel or the inter-robot guidance term drops the success rate from 100\% (full LHD) to 5\% and 19\% respectively. This confirms that both components are essential for robust multi-robot coordination.

\subsection{Scalability Analysis}
To evaluate the scalability of LHD for larger robot teams, we conducted additional experiments with 15, 20, 25, and 30 robots. 
Since LHD learns a gradient field from language-specified goals, it naturally generalizes to arbitrary team sizes without retraining. 
Using a model trained with 9 goals, we evaluated these expanded teams by assigning multiple agents to shared instructions.
As presented in Table \ref{tab:scalability}, LHD scales effectively for up to 20 robots, maintaining success rates above 86\% across all maps.
While performance remains robust in the Drop-Region and Conveyor maps even with 30 robots ($>$85\%), it gradually degrades in more constrained environments.
These results suggest that LHD provides a viable solution for scaling to larger robot teams, though its effectiveness may degrade in more constrained environments.

\subsection{Out-of-Distribution Generalization}
A key advantage of LHD is its ability to generalize to OOD scenarios without additional fine-tuning. 
We evaluate this capability using unseen obstacle layouts with two identical goals (e.g., apples), only one of which is reachable.
As baseline methods rely on Lang-SAM without assessing reachability, they blindly navigate toward the unreachable goal and inevitably collide with obstacles.
In contrast, LHD steers probability mass away from obstacles by setting $K(x) = 0$, thereby excluding blocked regions from the perturbed distribution $p_t(x)$.
This mechanism inherently filters out unreachable goals, ensuring that the reverse diffusion process generates trajectories toward the accessible target.

\begin{figure}[t]
    \centering 
    \includegraphics[width=1.0\linewidth]{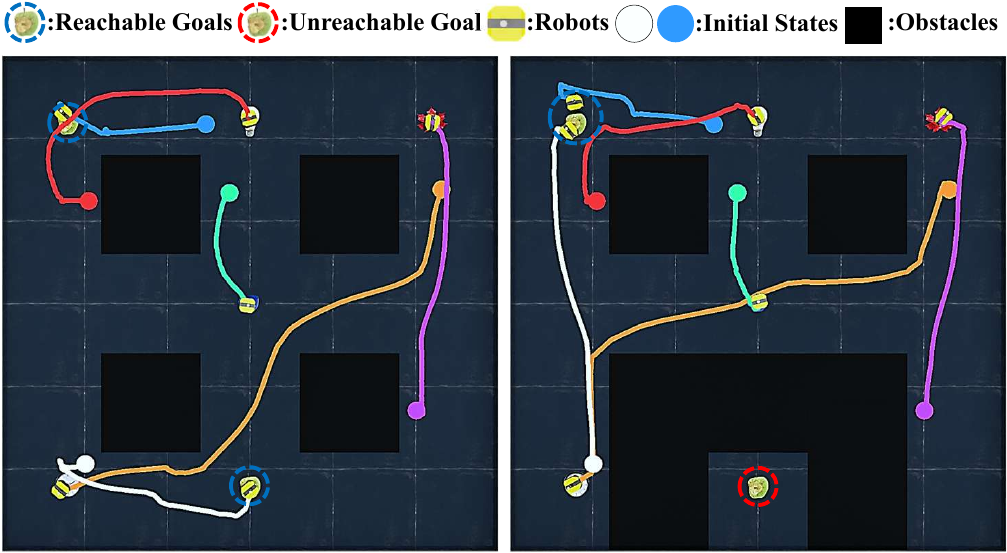}
    \caption{\textbf{Qualitative visualization of OOD generalization} (Left) In the unobstructed case, robots naturally navigate to their nearest targets. (Right) When one goal is unreachable, the robot autonomously redirects to the accessible target.}
    \label{fig:ood} 
\end{figure}

\begin{figure}[t]
    \centering
    \includegraphics[width=1.0\linewidth]{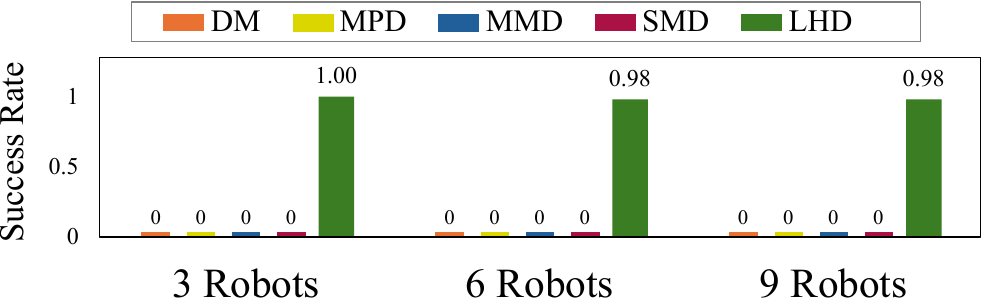}
    \caption{\textbf{OOD generalization performance.} Success rates averaged over 50 trials across varying team sizes (N=3,6,9) with unreachable goals.}
    \label{fig:ood_chart}
\end{figure}

This capability is qualitatively and quantitatively validated in Fig. \ref{fig:ood} and Fig. \ref{fig:ood_chart}, respectively. 
As visualized in Fig. \ref{fig:ood}, LHD autonomously redirects to accessible alternative when initial target are blocked.
Quantitatively, as shown in Fig. \ref{fig:ood_chart}, baseline methods achieve zero success rates in this scenario as they consistently attempt to navigate to the unreachable goal. LHD, however, maintains near-perfect success rates. This demonstrates that robust reachability awareness can arise from heat-inspired physical priors, without additional goal verification mechanisms.
\begin{figure}[t]
    \centering \includegraphics[width=1.0\linewidth]{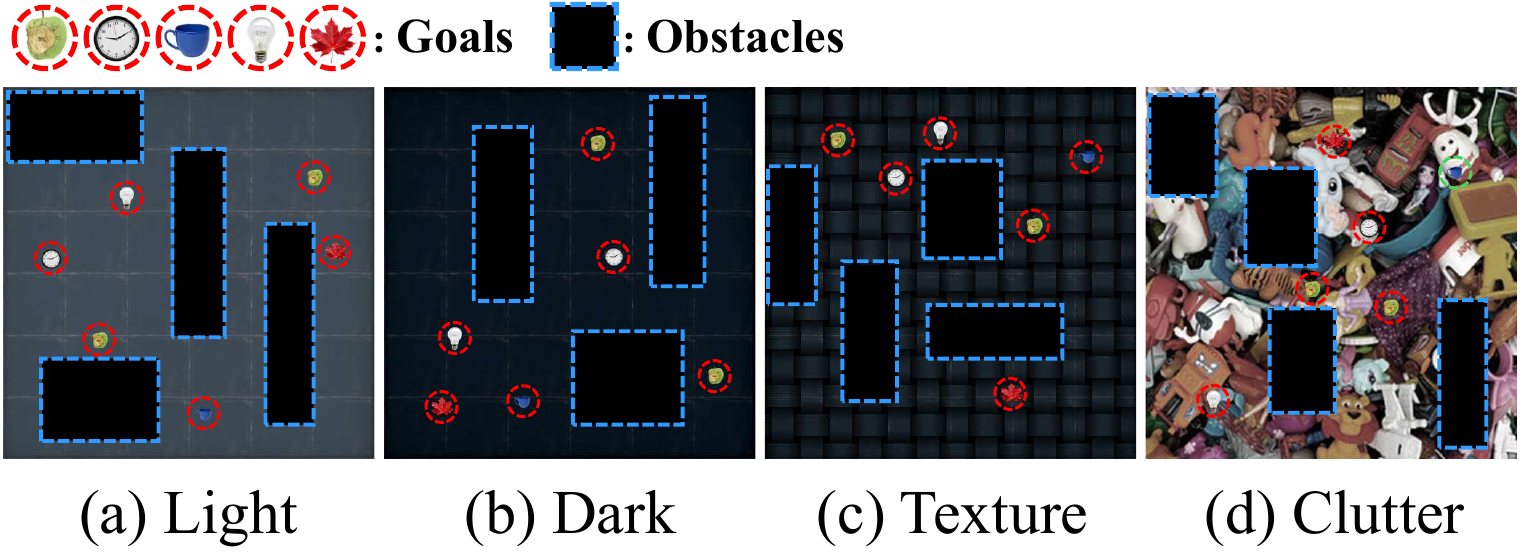}
    \caption{\textbf{Representative Snapshots of Visual Domain Shifts.} The environments feature lighting changes (Light, Dark), texture variations (Texture), and background clutter (Clutter).}
    \label{fig:visual_domain} 
    \vspace{-6pt}
\end{figure}
\begin{table}[t]
    \centering
    \caption{\textbf{Success Rate across Visual Domain Shifts in Highly Unstructured Environments.} The bottom row demonstrates the impact of visual data augmentation.}
    \label{tab:merged_results}
    \renewcommand{\arraystretch}{1.2}
    
    \resizebox{\columnwidth}{!}{%
        \begin{tabular}{l ccccc} 
            \toprule
            \textbf{Setting} & \textbf{Original} & \textbf{Light} & \textbf{Dark} & \textbf{Texture} & \textbf{Clutter} \\
            \midrule
            3 Robots (w/o Aug.) & 1 & 0.98 & 1 & 0.98 & 0 \\
            6 Robots (w/o Aug.) & 0.86 & 0.06 & 0.84 & 0.62 & 0 \\
            9 Robots (w/o Aug.) & 0.82 & 0 & 0 & 0 & 0 \\
            \midrule
            \textbf{9 Robots (w/ Aug.)} & \textbf{0.84} & \textbf{0.86} & \textbf{0.86} & \textbf{0.84} & 0 \\
            \bottomrule
        \end{tabular}%
    }
\end{table}
\begin{figure}[!ht]
    \centering
    \includegraphics[width=1.0\linewidth]{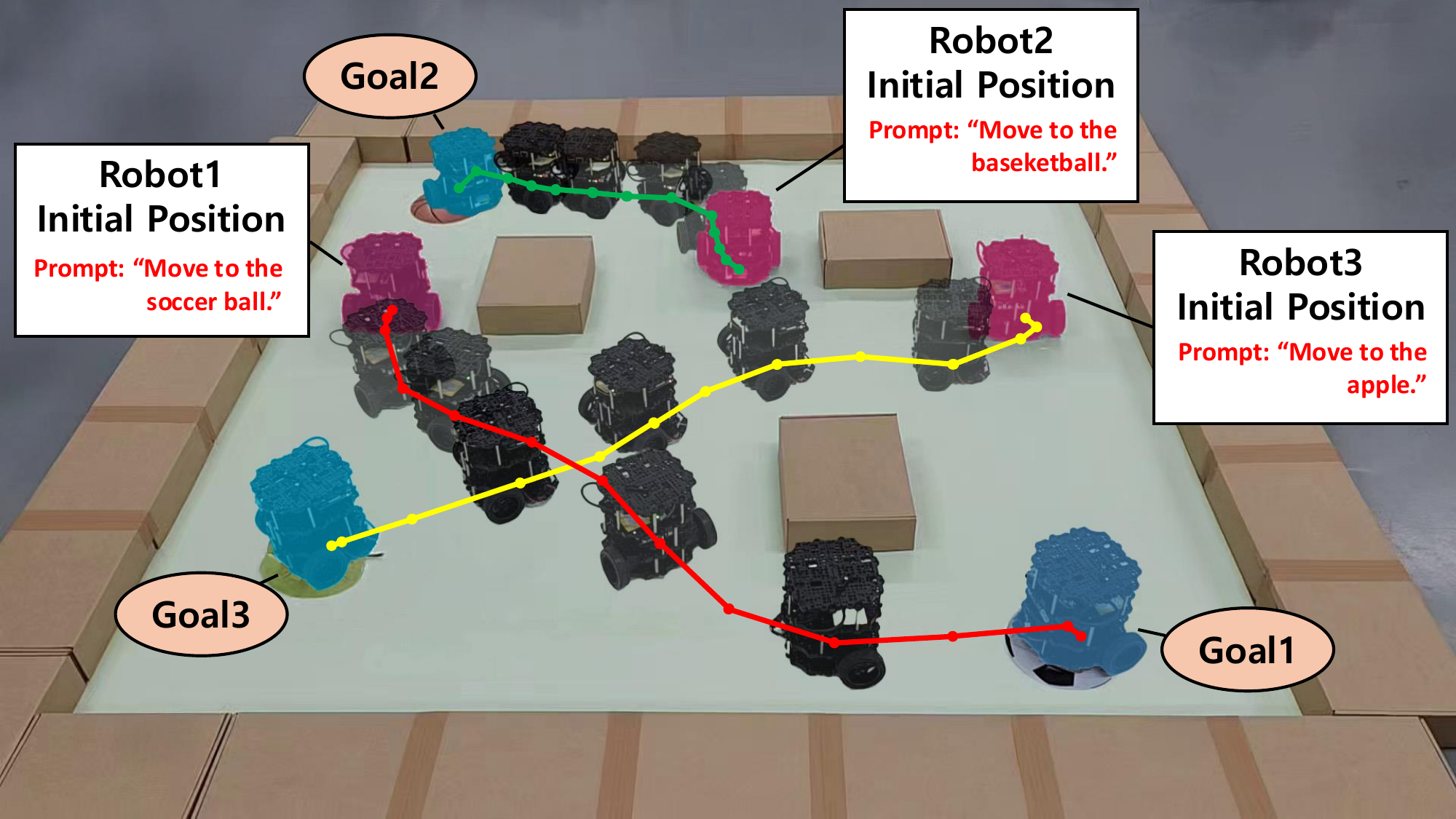}
    \caption{\textbf{Real-world validation.} The colored lines (red, yellow, and green) represent the actual executed trajectories of three robots navigating to their respective goals.}
    \label{fig:real_top_down}
\end{figure}
\subsection{Robustness to Visual Domain Shifts}
We evaluate the zero-shot performance of LHD across four unseen visual domains—\textit{Light}, \textit{Dark}, \textit{Texture}, and \textit{Clutter}—in \textbf{50 highly unstructured environments} (representative snapshots in Fig. \ref{fig:visual_domain}).
While the model generalizes for 3 robots across most domains except \textit{Clutter} settings, success rates drop to zero for 9 robots across all unseen domains (Table \ref{tab:merged_results}). 
To mitigate this, we augmented the original background with randomized brightness, contrast, Gaussian noise, and occasional grayscale/inversion. 
As demonstrated in the last row of Table \ref{tab:merged_results}, while the Clutter domain remains challenging due to its visually complex background with irregular objects, this approach significantly restores success rates for large teams (9 robots) in the \textit{Light}, \textit{Dark}, and \textit{Texture} domains. 
These findings suggest that visual augmentation facilitates robustness to novel, unstructured settings. 
Consequently, training on diverse real-world backgrounds could further extend generalization to more extreme visual domain shifts.

\subsection{Real-world Experiments}
To validate LHD's practical applicability, we deploy our method on three TurtleBot3 robots in the real world, comparing it against the strongest baseline, SMD.

\textbf{Setup.} We conduct 20 test cases with randomized start and goal positions using an Intel RealSense L515 camera positioned overhead to capture a top-down view image. Notably, the raw RGB image is directly fed to our model without any extrinsic calibration or preprocessing. This setup demonstrates that LHD can operate with off-the-shelf RGB cameras, including smartphone cameras.

\textbf{Results.} Table \ref{tab:real_world_result} summarizes the real-world performance. Both LHD and SMD achieved a success rate of 18/20 (90\%). 
Despite the comparable success rates, a substantial disparity exists in computational efficiency. LHD generates solutions in an average of 0.58~seconds. In contrast, SMD requires 45.87~seconds on average to converge. This represents an approximately 80$\times$ speedup, highlighting LHD's suitability for time-sensitive real-world applications compared to SMD.

\section{CONCLUSIONS}

In this work, we introduce Language-conditioned Heat-inspired Diffusion (LHD) for multi-robot motion planning from language instructions and visual inputs. By integrating a collision-avoiding diffusion kernel, LHD injects an explicit collision-avoidance inductive bias into the language-guided diffusion process, improving robustness to unseen scenarios. 
The limitation of our approach is the initial transient of the heat propagation can cause early boundary deviation, leading to slightly inefficient trajectories (Sec.~\ref{comparison}). 
Future work could explore utilizing LHD as a Flow Matching-based ~\cite{lipman2022flow}  local replanner for real-time closed-loop replanning against dynamic obstacles. 
Additionally, extending to 3D heat transfer for high-DoF systems like multiple mobile-manipulators or humanoids could be investigated.

\begin{table}[t]
\centering
\caption{\textbf{Real-world performance.} Comparison of success rates and planning times between LHD and SMD.}
\label{tab:real_world_result}
\small 
\setlength{\tabcolsep}{7.0pt} 
\begin{tabular}{l|cc}
\toprule
Method & Success Rate & Average Planning Time (s)\\
\midrule
SMD & \textbf{18 / 20} & 45.87 \\
\textbf{LHD (Ours)} & \textbf{18 / 20} & \textbf{0.58} \\
\bottomrule
\end{tabular}
\end{table}

\bibliographystyle{IEEEtran}
\bibliography{reference}

\end{document}